\documentclass[letterpaper, 10pt, conference]{ieeeconf}      

\IEEEoverridecommandlockouts                              
\usepackage{amsmath} 
\usepackage{amssymb}  
\usepackage{cite}
\usepackage{threeparttable}
\usepackage{stfloats}
\usepackage{comment}
\usepackage{gensymb}
\usepackage{subfig}
\usepackage{booktabs} 
\usepackage{hyperref}

\title{\LARGE \bf
History Encoding Representation Design for Human Intention Inference
}

\author{
       Zhuo Xu,
       and Masayoshi Tomizuka\\
       University of California, Berkeley
}

\usepackage{algorithm,algpseudocode}
\usepackage[algo2e]{algorithm2e} 
\usepackage{graphicx}\usepackage{multirow}
\usepackage{url}
\usepackage{color,soul}
\usepackage{bm}
\begin{document}

\maketitle
\thispagestyle{empty}
\pagestyle{empty}

\begin{abstract}
In this extended abstract, we investigate the design of learning representation for human intention inference. In our designed human intention prediction task, we propose a history encoding representation that is both interpretable and effective for prediction. Through extensive experiments, we show our prediction framework with a history encoding representation design is successful on the human intention prediction problem.
\end{abstract}

\section{Introduction and Setup}

\begin{figure}[b]
\centering
\includegraphics[width=0.9\linewidth]{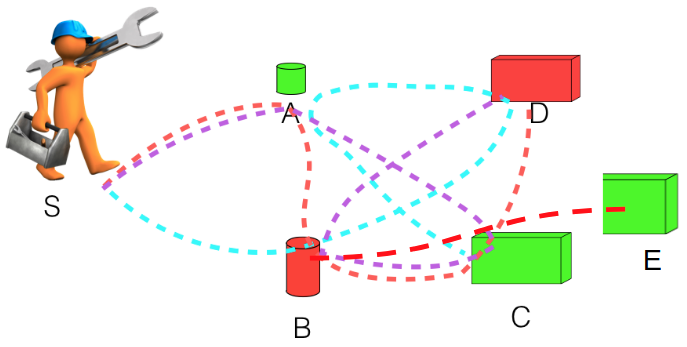}
\caption{Overview of the multi-step pick and place task.}
\label{fig:bp_overview}
\end{figure}

In this extended abstract, we investigate the selection of history encoding representation for the application of human behavior prediction in a multi-step pick and place task based on supervised learning. Prior works have shown that the reasoning of representations can significantly improve learning-based inference performance \cite{xu2018zero, tang2019disturbance, xu2017cascade, retinagan, xu2019toward, xu2020cocoi, xu2020guided, chen2020end, chang2020cascade}. We design a novel and interpretable representation formulation to effectively recognize the scene and encode the history information so as to obtain a fast and accurate human behavior prediction module. The task is designed as follows (overview shown in Fig. \ref{fig:bp_overview}): A human worker need to pick up two objects A, and B. The worker shall put object A onto either one of the C and E pads, and object B onto pad D. The worker has to perform each pick, place and transport operation one by one, resulting in a sequential operation strategies. We can observe the human performing the task from a third-party-view camera image stream, and the goal of the task is to infer the strategy as soon and accurately as possible.

\begin{figure}
\minipage{0.09\textwidth}
  \includegraphics[width=\linewidth]{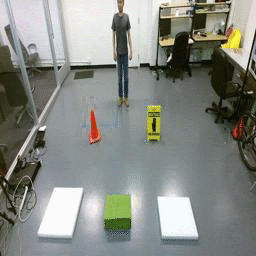}
\endminipage\hfill
\minipage{0.09\textwidth}
  \includegraphics[width=\linewidth]{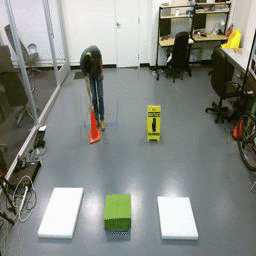}
\endminipage\hfill
\minipage{0.09\textwidth}%
  \includegraphics[width=\linewidth]{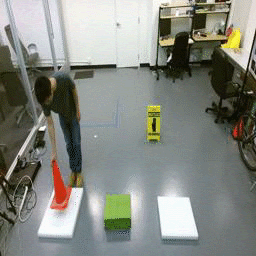}
\endminipage
\minipage{0.09\textwidth}%
  \includegraphics[width=\linewidth]{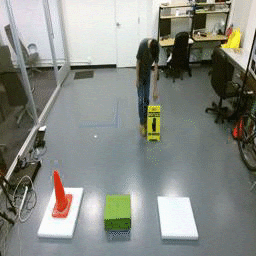}
\endminipage
\minipage{0.09\textwidth}%
  \includegraphics[width=\linewidth]{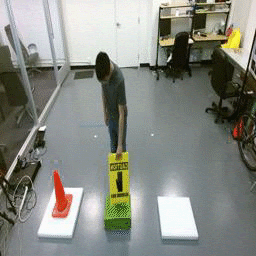}
\endminipage
\caption{One strategy the human worker takes to complete the multi-step pick and place task. The human worker first pick the red object and put it at the left white pad, and then pick the yellow object and put it at the green pad}
\label{fig:bp_strategy}
\end{figure}

\begin{figure}
\minipage{0.09\textwidth}
  \includegraphics[width=\linewidth]{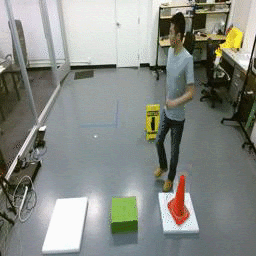}
\endminipage\hfill
\minipage{0.09\textwidth}
  \includegraphics[width=\linewidth]{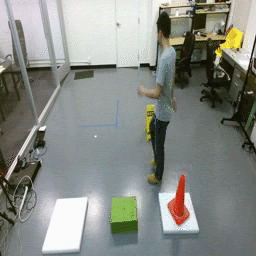}
\endminipage\hfill
\minipage{0.09\textwidth}%
  \includegraphics[width=\linewidth]{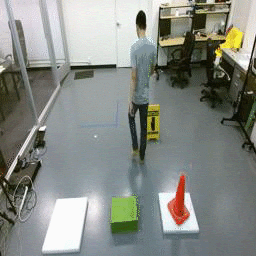}
\endminipage
\minipage{0.09\textwidth}%
  \includegraphics[width=\linewidth]{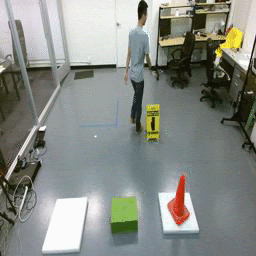}
\endminipage
\minipage{0.09\textwidth}%
  \includegraphics[width=\linewidth]{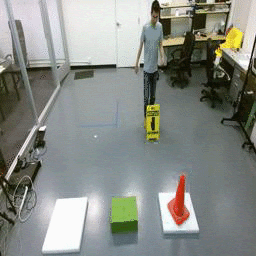}
\endminipage
\caption{One sub-task the human worker performs: transport from the right white pad to the yellow object.}
\label{fig:bp_step1}
\end{figure}

\begin{figure}
\minipage{0.09\textwidth}
  \includegraphics[width=\linewidth]{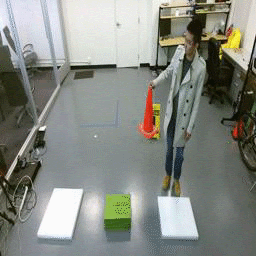}
\endminipage\hfill
\minipage{0.09\textwidth}
  \includegraphics[width=\linewidth]{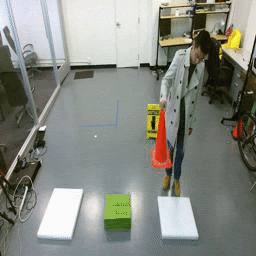}
\endminipage\hfill
\minipage{0.09\textwidth}%
  \includegraphics[width=\linewidth]{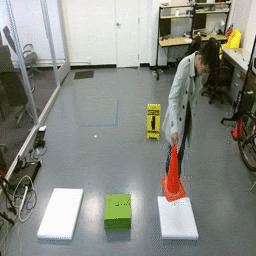}
\endminipage
\minipage{0.09\textwidth}%
  \includegraphics[width=\linewidth]{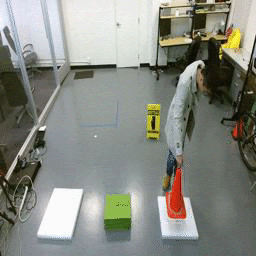}
\endminipage
\minipage{0.09\textwidth}%
  \includegraphics[width=\linewidth]{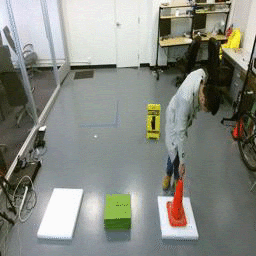}
\endminipage
\caption{One sub-task the human worker performs: place the red object onto the right white pad.}
\label{fig:bp_step2}
\end{figure}

To study the behavior prediction methodology, We collect a large video dataset of a human volunteer performing the task, as shown in Fig. \ref{fig:bp_strategy}. To finish this task, the human worker has 12 strategies in total, Fig. \ref{fig:bp_strategy} shows one of them. We decompose the whole task into a series of subtasks, such as transporting from one location to another, picking and placing an object. There are in total 22 such sub-tasks, and Fig. \ref{fig:bp_step1} and Fig. \ref{fig:bp_step2} shows two of them.

\section{Behavior Prediction Framework and History Encoding Representation}

\begin{figure}
\centering
\includegraphics[width=\linewidth]{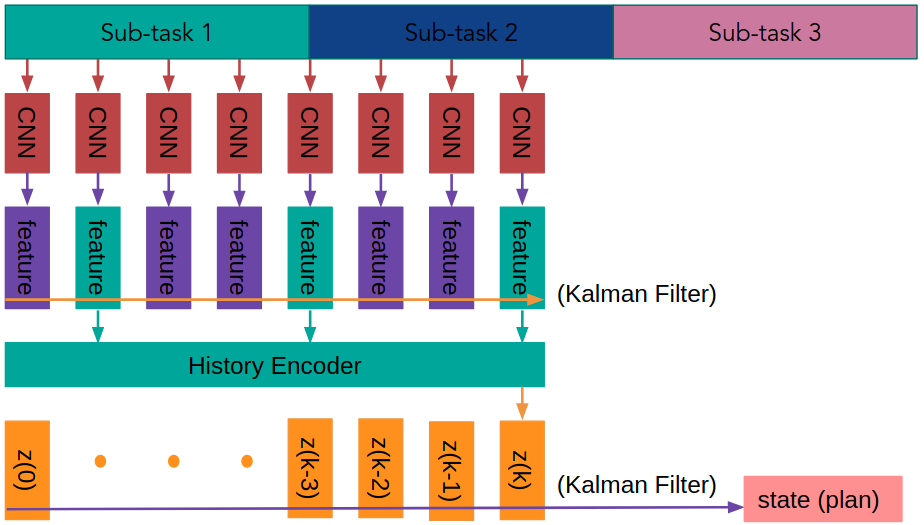}
\caption{The two stage behavior prediction framework.}
\label{fig:bp_framework}
\end{figure}

We propose a two stage behavior prediction framework that is shown in Fig. \ref{fig:bp_framework}. First, the camera image is fed into a VGG convolutional neural network \cite{simonyan2014very} classifier to extract a feature for the current operation. We choose the features to be the sub-task that the human is operating at the current frame, resulting in a 22 dimensional one-hot vector encoding the 22 possible sub-task at the current video frame. We then run a steady state Kalman filter to smooth the feature vectors from the CNN. Then the feature history is encoded into a history encoding representation, which is defined as follows.

\begin{figure}
\centering
\includegraphics[width=\linewidth]{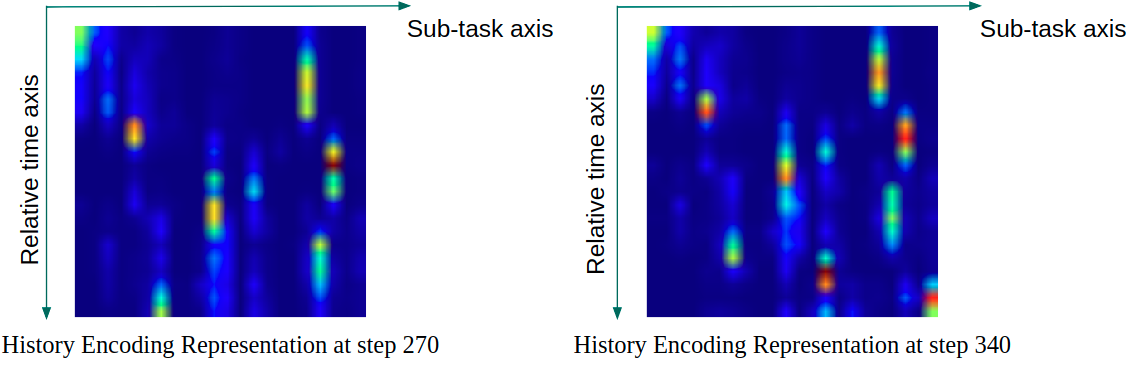}
\caption{Two history encoding representation obtained in one episode at two different time steps.}
\label{fig:bp_history_representation}
\end{figure}

The history encoding representation uniformly (sampling time length can be different) select a fixed number of history features to form the information vector. The fixed number of history features is chosen to be 22, which is the same as the the dimension of the feature vector. Therefore, the history encoding representations can be rendered using 22 by 22 square figures, as shown in Fig. \ref{fig:bp_history_representation}. In Fig. \ref{fig:bp_history_representation}, the history encoding representations can be interpreted according to a relative time axis and a sub-task axis. Concretely, for each row vector with same relative time, the heat map represents the probability that the human worker is performing one of the sub-tasks at the relative time. Finally, this history encoding representation is fed into a final neural network to produce the strategy that the human is taking with ground truth strategy label supervision. After acquiring a series of estimations, we run a steady state Kalman filter to calculate the posterior state estimate. The initial state of the Kalman filter is assumed to be a discrete uniform distribution over all 12 plans.

\section{Performance and Discussions}

We collect a large dataset of 50987 frames of 256 by 256 rgb images recording a human volunteer performing the task. Sample images can be found in Fig. \ref{fig:bp_strategy}, \ref{fig:bp_step1}, and \ref{fig:bp_step2}. We randomly separate the dataset into a training set of 46737 images and 4250 testing images. We finetune a VGG classification network pre-trained on ImageNet \cite{deng2009imagenet} (with the first 4 convolutional layers fixed) on the training set (with data augmentation of randomly cropping the images to size of 224 by 224), and evaluated the results on the testing set. We obtain a classification accuracy on the test set of 51\% top 1 accuracy, and 75\% and 93\% top 2 and top 4 accuracy. The history encoding representations shown in Fig. \ref{fig:bp_history_representation} also show clear and contingent sub-task classification performance.

\begin{figure}
\minipage{0.2\textwidth}
  \includegraphics[width=\linewidth]{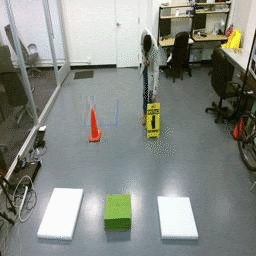}
\endminipage\hfill
\minipage{0.2\textwidth}
  \includegraphics[width=\linewidth]{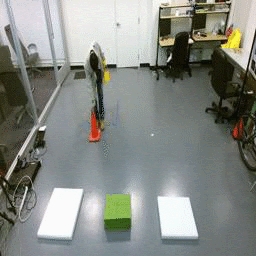}
\endminipage\hfill
\newline
\minipage{0.2\textwidth}
  \includegraphics[width=\linewidth]{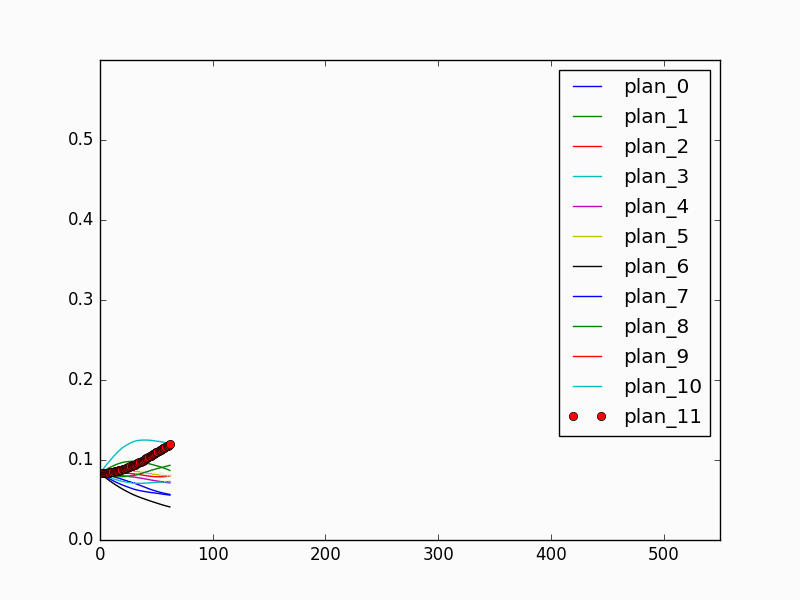}
\endminipage\hfill
\minipage{0.2\textwidth}
  \includegraphics[width=\linewidth]{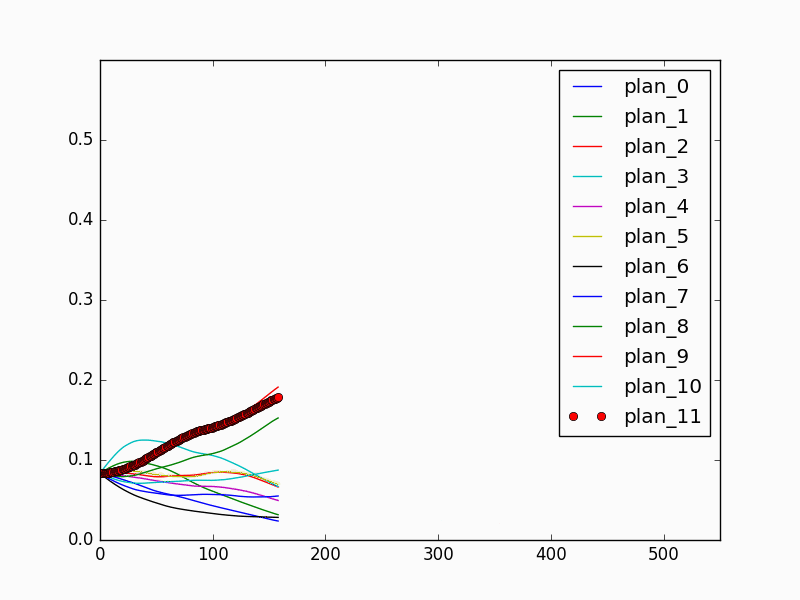}
\endminipage\hfill
\caption{First half of the behavior prediction model performance in an online episode.}
\label{fig:bp_half1}
\end{figure}

\begin{figure}
\minipage{0.2\textwidth}%
  \includegraphics[width=\linewidth]{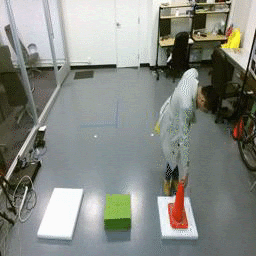}
\endminipage
\minipage{0.2\textwidth}%
  \includegraphics[width=\linewidth]{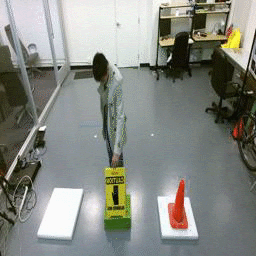}
\endminipage
\newline
\minipage{0.2\textwidth}%
  \includegraphics[width=\linewidth]{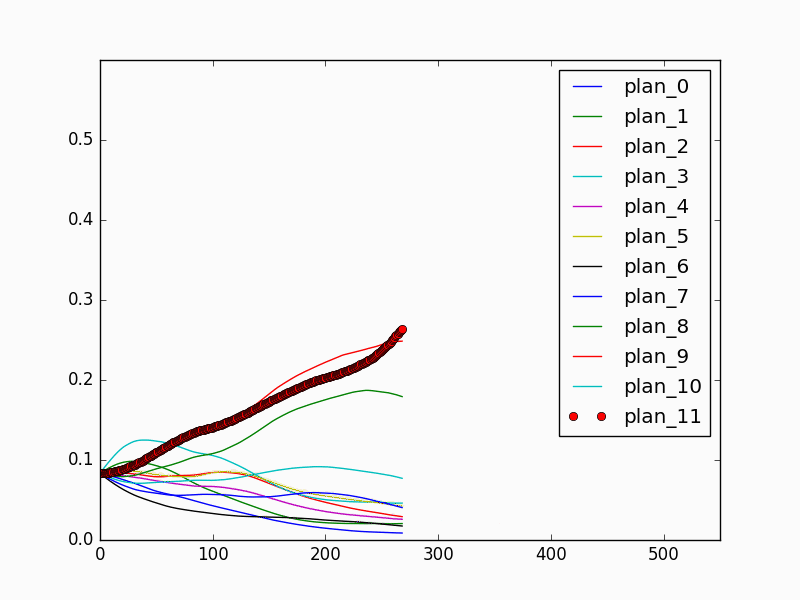}
\endminipage
\minipage{0.2\textwidth}%
  \includegraphics[width=\linewidth]{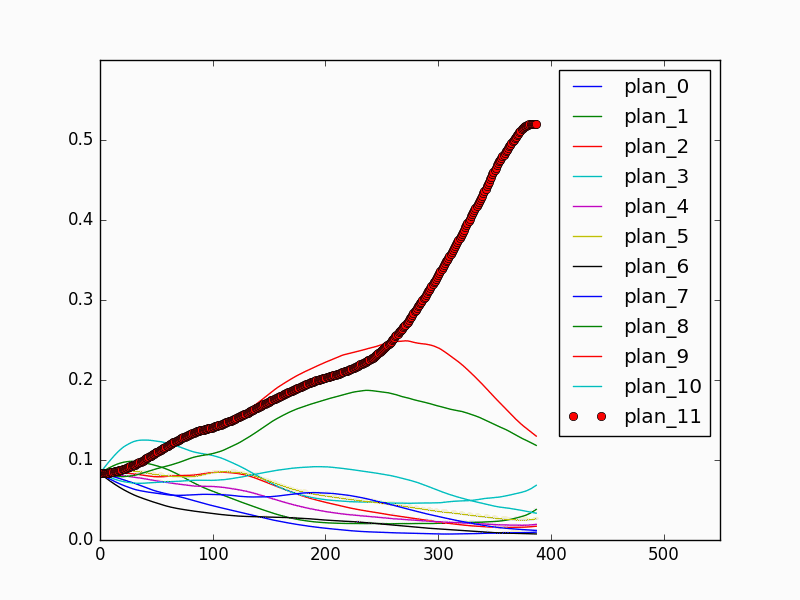}
\endminipage
\caption{Second half of the behavior prediction model performance in an online episode.}
\label{fig:bp_half2}
\end{figure}

The fully connected network that maps from the history encoding representation to the strategy estimation is trained using 12 episodes (1 for each strategy) with 4250 frames of images, and tested on another 12 episodes (1 for each strategy) with 4047 frames. We result with a 32.4\% top 1 classification accuracy and a 85.6\% top 5 classification accuracy. The relatively low classification accuracy is due to that in the early stage of the episode, it is not possible to identify the correct strategy. For prediction performance, the model achieves 100\% final classification success rate. During online prediction, the model can achieve human-level performance by converging quickly to the correct strategy once enough evidence is observed. Fig \ref{fig:bp_half1} and Fig. \ref{fig:bp_half2} show the behavior prediction model performance in an online episode. It is shown that in the first half of the episode (Fig \ref{fig:bp_half1}) when the human volunteer picks the red object and the yellow object, the prediction model cannot determine which object the human is placing first, or which strategy the human is taking. Therefore, the thick red probability curve which corresponds to the correct strategy is similarly as high as the thin red curve, which corresponds to a wrong but indistinguishable strategy. In the second half of the episode (Fig \ref{fig:bp_half2}), as the human approaches the right white pad and places the red object, the model is able to correctly identify the strategy, and the thick red curve rises above the thin red curve, and finally reaches to far above all other curves. The experiments show that the history encoding representation and the prediction model are able to capture the history information and achieve satisfying prediction performance.


\bibliography{references}
\bibliographystyle{ieeetr}
\end{document}